\relax
\documentclass[letterpaper]{article}
\usepackage[preprint]{neurips_2021} 
\usepackage[utf8]{inputenc} 
\usepackage[T1]{fontenc}    
\usepackage{hyperref}       
\usepackage{url}            
\usepackage{booktabs}       
\usepackage{amsfonts}       
\usepackage{nicefrac}       
\usepackage{microtype}      
\usepackage{xcolor}         
\usepackage{algpseudocode}
\usepackage{algorithm}
\usepackage{graphicx}
\usepackage{amsfonts}
\usepackage{amsmath}

\usepackage{booktabs}
\usepackage{multirow}
\title{Wasserstein Unsupervised Reinforcement Learning}
\author{%
    Shuncheng He\\
    Department of Automation\\
    Tsinghua University\\
    Beijing, China\\
    \texttt{hesc16@mails.tsinghua.edu.cn} \\
    \And
    Yuhang Jiang\\
    Department of Automation\\
    Tsinghua University\\
    Beijing, China\\
    \texttt{jiangyh19@mails.tsinghua.edu.cn} \\
    \AND
    Hongchang Zhang\\
    Department of Automation\\
    Tsinghua University\\
    Beijing, China\\
    \texttt{hc-zhang19@mails.tsinghua.edu.cn} \\
    \And
    Jianzhun Shao\\
    Department of Automation\\
    Tsinghua University\\
    Beijing, China\\
    \texttt{sjz18@mails.tsinghua.edu.cn} \\
    \AND
    Xiangyang Ji\\
    Department of Automation\\
    Tsinghua University\\
    Beijing, China\\
    \texttt{xyji@mail.tsinghua.edu.cn} \\  
}

\begin{document}

\maketitle
\begin{abstract}
    Unsupervised reinforcement learning aims to train agents to learn a handful of policies or skills in environments without external reward. These pre-trained policies can accelerate learning when endowed with external reward, and can also be used as primitive options in hierarchical reinforcement learning. Conventional approaches of unsupervised skill discovery feed a latent variable to the agent and shed its empowerment on agent’s behavior by \textit{mutual information} (MI) maximization. However, the policies learned by MI-based methods cannot sufficiently explore the state space, despite they can be successfully identified from each other. Therefore we propose a new framework Wasserstein unsupervised reinforcement learning (WURL) where we directly maximize the distance of state distributions induced by different policies. Additionally, we overcome difficulties in simultaneously training $N(N>2)$ policies, and amortizing the overall reward to each step. Experiments show policies learned by our approach outperform MI-based methods on the metric of Wasserstein distance while keeping high discriminability. Furthermore, the agents trained by WURL can sufficiently explore the state space in mazes and MuJoCo tasks and the pre-trained policies can be applied to downstream tasks by hierarchical learning.
\end{abstract}
\section{Introduction}
Autonomous agents can learn to solve challenging tasks by deep reinforcement learning, including locomotive manipulation (\cite{lillicrap2015continuous}; \cite{haarnoja2018soft}) and game playing (\cite{mnih2015human}; \cite{silver2016mastering}). The reward signal specified by the task plays an important role of supervision in reinforcement learning. However, recent research reveals the possibilities that agents can acquire diverse \textit{skills} or \textit{policies} in the absence of reward signal (\cite{eysenbach2018diversity}; \cite{gregor2016variational}; \cite{achiam2018variational}). This setting is called \textit{unsupervised reinforcement learning}.\par
Practical applications of unsupervised reinforcement learning have been studied. The skills learned without reward can serve as primitive options for hierarchical RL in long horizon tasks (\cite{eysenbach2018diversity}). Also the primitive options may be useful for transferring across different tasks. In model-based RL, the learned skills enables the agent to plan in the skill space (\cite{sharma2019dynamics}). Unsupervised learning methods may alleviate the cost of supervision: in certain cases, designing reward function requires human supervision (\cite{NIPS2017_d5e2c0ad}). The intrinsic reward derived from unsupervised learning can enhance exploration when combined with task reward (\cite{houthooft2016vime}; \cite{gupta2018meta}).\par
The key point of unsupervised reinforcement learning is how to learn a set of policies that can sufficiently explore the state space. Previous methods make use of a latent variable and maximize the mutual information (MI) between the latent variable and the behavior (\cite{eysenbach2018diversity}). Consequently the diversity in the latent space is cast into the state space. These methods are able to obtain different skills which are distinguishable from each other. However, limitations of MI-based methods are pointed out as the diversity of learned skills is restricted by the Shannon entropy of the latent variable. In addition, discriminability of skills does not always lead to the goal of sufficient exploration of the environment.\par
In this paper, we propose a new approach of unsupervised reinforcement learning which is essentially different form MI-based methods. The motivation of our method is to increase the discrepancy of learned policies so that the agents can explore the state space extensively and reach the state as ``far'' as possible compared to other policies. This idea incentivizes us to employ a \textit{geometry-aware} metric to measure the discrepancy between the state distributions induced by different policies. In recent literature of generative modeling, the optimal transport (OT) cost is a new direction to measure distribution distance (\cite{tolstikhin2018wasserstein}) since it provides a more geometry-aware topology than $f$-divergences, including GAN (\cite{NIPS2016_cedebb6e}). Therefore, we choose Wasserstein distance, a well-studied distance from optimal transport, to measure the distance between different policies in unsupervised reinforcement learning. By maximizing Wasserstein distance, the agents equipped with different policies may drive themselves to enter different areas of state space and keep as ``far'' as possible from each other to earn greater diversity.\par
Our contributions are four-fold. First, we propose a novel framework adopting Wasserstein distance as discrepancy measure for unsupervised reinforcement learning. This framework is well-designed to be compatible with various Wasserstein distance estimation algorithms, both in primal form and in dual form. Second, as Wasserstein distance is defined on \textbf{two} distributions, we extend our framework to multiple policy learning. Third, to address the problem of sparse reward provoked by Wasserstein distance estimation, we devise an algorithm to amortize Wasserstein distance between two bunches of samples to stepwise intrinsic reward. Four, we empirically demonstrate our approach surpasses the diversity of MI-based methods and can cover the state space by incremental learning.
\begin{figure*}
    \centering
    \includegraphics[width=0.9\textwidth]{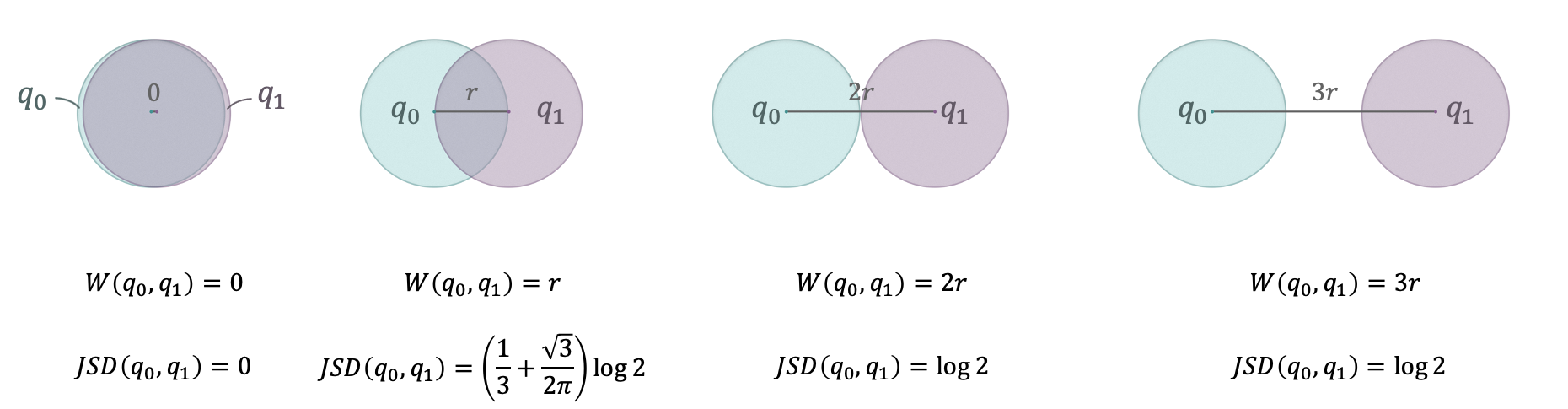}
    \caption{Examples of Jensen-Shannon divergence and Wasserstein distance between $q_{0}$ and $q_{1}$. $q_{0}$ and $q_{1}$ are uniform distributions over a round plate with radius $r$. The distance between the centers of $q_{0}, q_{1}$ varies from 0 to $3r$.}
    \label{fig:1}
\end{figure*}
\section{Distribution discrepancy measure}
In this Section, we briefly review Wasserstein distance and its estimation methods.
\subsection{Wasserstein distance and optimal transport}
Measuring discrepancy or distance between two probability distributions can be seen as a transport problem (\cite{villani2009optimal}). Consider two distributions $p,q$ on domains $\mathcal{X}\subseteq\mathbb{R}^{n}$ and $\mathcal{Y}\subseteq\mathbb{R}^{m}$. Let $\Gamma[p,q]$ be the set of all distributions on product space $\mathcal{X}\times\mathcal{Y}$, with their marginal distributions on $\mathcal{X}$ and $\mathcal{Y}$ being $p,q$ respectively. Therefore, given a proper cost function $c(x,y):\mathcal{X}\times\mathcal{Y}\rightarrow\mathbb{R}$ for moving mass from $x$ to $y$, the Wasserstein distance is defined as
\begin{equation}
    W_{c}(p,q)=\inf_{\gamma\in\Gamma[p,q]}\int_{\mathcal{X}\times\mathcal{Y}}c(x,y)d\gamma.
\end{equation}
The joint distribution family $\Gamma[p,q]$ essentially forms a family of bijective plans transporting probability mass from $p$ to $q$. Minimizing the transport cost is an optimal transport problem. The optimization problem suffers from supercubic complexity, since the problem becomes linear programming when $\mathcal{X}$ and $\mathcal{Y}$ are finite discrete sets (\cite{cuturi2013sinkhorn}; \cite{NIPS2016_2a27b814}). To avoid suboptimal solutions, a regularizer is added to the optimization objective. The smoothed Wasserstein distance is defined as
\begin{equation}
    \tilde{W}_{c}(p,q)=\inf_{\gamma\in\Gamma[p,q]}\left[\int_{\mathcal{X}\times\mathcal{Y}}c(x,y)d\gamma+\beta KL(\gamma|pq)\right].
\end{equation}
Minimizing the cost together with the KL divergence encourages the joint distribution $\gamma(x,y)$ to move close to $p(x)q(y)$. As $\beta\rightarrow 0$, the smoothed distance converges to $W_{c}(p,q)$ (\cite{pacchiano2020learning}).\par
The objective function is convex if $c(x,y)$ is a proper cost function. Therefore the infimum can be calculated either by primal formulation or dual formulation. In Section 2.2 and 2.3, we introduce practical methods estimating Wasserstein distance from distribution samples.
\subsection{Primal form estimation}
Solving the optimal transport problem from the primal formulation is hard. However, the problem has analytical solutions when the distributions are on one-dimensional Euclidean space and cost function is $l_{p}$ measure ($p\geq 0$) (\cite{rowland2019orthogonal}). Inspired by 1-D Wasserstein distance estimation, we may estimate Wasserstein distance in high dimensional Euclidean spaces by projecting distributions to $\mathbb{R}$. Suppose $p,q$ are probability distributions on $\mathbb{R}^{d}$. For a vector $v$ on the unit sphere $S^{d-1}$ in $\mathbb{R}^{d}$, the projected distribution $\Pi_{v}(p)$ is the marginal distribution along the vector by integrating $p$ in the orthogonal space of $v$. Estimating Wasserstein distance on 1-D space results sliced Wasserstein distance (SWD) (\cite{wu2019sliced}; \cite{kolouri2018sliced}):
\begin{equation}
    SW(p,q)=\mathbb{E}_{v\sim U(S^{d-1})}\left[W(\Pi_{v}(p),\Pi_{v}(q))\right],
\end{equation}
where $U(S^{d-1})$ means the uniform distribution on unit sphere $S^{d-1}$. In practical use, the projected distribution $\Pi_{v}(\hat{p})$ of empirical distribution $\hat{p}=\frac{1}{N}\sum_{n=1}^{N}\delta_{x_{n}}$ can be written as $\Pi_{v}(\hat{p})=\frac{1}{N}\sum_{n=1}^{N}\delta_{\langle x_{n},v\rangle}$,
where $\langle\cdot,\cdot\rangle$ denotes inner product and $\delta$ is Dirac distribution.\par
To reduce estimation bias of SWD, \citet{rowland2019orthogonal} proposed projected Wasserstein distance (PWD) by disentangling coupling calculation and cost calculation. PWD obtains optimal coupling by projecting samples to $\mathbb{R}$, and calculates costs in original space $\mathbb{R}^{d}$ rather than the projected space.
\subsection{Dual form estimation}
Define set $\mathcal{A}=\{(u,v)|\forall(x,y)\in\mathcal{X}\times\mathcal{Y}:u(x)-v(y)\leq c(x,y)\}$. By Fenchel-Rockafellar duality, the dual form of Wasserstein distance is (\cite{villani2009optimal})
\begin{equation}
    W_{c}(p,q)=\sup_{(\mu,\nu)\in\mathcal{A}}\mathbb{E}_{x\sim p(x),y\sim q(y)}\left[\mu(x)-\nu(y)\right],
\end{equation}
where $\mu:\mathcal{X}\rightarrow\mathbb{R}$ and $\nu:\mathcal{Y}\rightarrow\mathbb{R}$ are continuous functions on their domains. The dual formulation provides us a neural approach to estimate Wasserstein distance, circumventing the difficulties to find the optimal transport plan between two probability distributions. The dual form of smoothed Wasserstein distance shows more convenience since there are no constraints on $\mu,\nu$:
\begin{equation}
    \tilde{W}_{c}(p,q)=\sup_{\mu,\nu}\mathbb{E}_{x\sim p(x),y\sim q(y)}\left[\mu(x)-\nu(y)-\beta\exp\left(\frac{\mu(x)-\nu(y)-c(x,y)}{\beta}\right)\right].
    \label{eqn:1}
\end{equation}
Alternative dual formulation emerges when $\mathcal{X}=\mathcal{Y}$. It is the most common case that two distributions are defined in the same space. Under this assumption, Kantorovich-Rubinstein duality gives another dual form objective in which only one function with Lipschitz constraint is optimized (\cite{villani2009optimal}).
\begin{equation}
    W_{c}(p,q)=\sup_{\|f\|_{L}\leq 1}\mathbb{E}_{x\sim p(x),y\sim q(y)}\left[f(x)-f(y)\right].
    \label{eqn:2}
\end{equation}
The maxima of dual problem theoretically equals to the minima of primal problem. However, sliced Wasserstein distance and projected Wasserstein distance have no such guarantee. Nonetheless, the primal form estimation methods show competitive accuracy empirically. 
\section{Wasserstein unsupervised reinforcement learning}
\subsection{MI-based unsupervised reinforcement learning}
Traditional unsupervised reinforcement learning adopts mutual information to seek diverse skills. Mutual information between two random variables is popularly perceived as the degree of empowerment (\cite{gregor2016variational}; \cite{kwon2021variational}): $I(X;Y)=H(X)-H(X|Y)=H(Y)-H(Y|X)$.
For instance, DIAYN (\cite{eysenbach2018diversity}) mainly aims to maximize $I(S;Z)$, the mutual information between latent variables and states reached by agent. Conventionally, the prior of latent variable $p(z)$ is fixed to uniform distribution which has maximal entropy. The maximization process of $I(S;Z)$ broadcasts the diversity in $Z$ to states $S$ through policy  $\pi(a|s,z)$. However, estimating mutual information involves intractable posterior distribution $p(z|s)$. With the tool of variational inference, a feasible lower bound is ready by approximating $p(z|s)$ with $q_{\phi}(z|s)$. We call $q_{\phi}(z|s)$ a learned discriminator trying to recognize the latent variable behind the policy from behavior. For example, the discriminator is a neural net predicting labels as in classification tasks when $p(z)$ is a categorical distribution.\par
From the view of optimization, the agent and the discriminator are trained in the cooperative way to maximize the same objective. This learning process comes to the end immediately after the discriminator can successfully infer $z$ behind the policy. However, the learned policy is not naturally diverse enough. We will explain this claim by a simple example. Suppose the latent variable $Z$ is randomly selected from $\{0,1\}$. The mutual information $I(S;Z)$ equals to the Jensen-Shannon divergence between the two conditional probability distributions $q_{0}=p(s|Z=0)$ and $q_{1}=p(s|Z=1)$. As illustrated in Fig. \ref{fig:1}, the JS divergence reaches the maximal value when the supports of $q_{0},q_{1}$ do not overlap. Actually, the decomposition of mutual information $I(S;Z)=H(Z)-H(Z|S)$ also implies that $I(S;Z)$ is upper bounded by $H(Z)$, which is fixed by a predetermined distribution.\par
To address this issue, we propose our method that uses Wasserstein distance as intrinsic reward to encourage the agent to explore for farther states. In Fig. \ref{fig:1}, Wasserstein distance provides information about how far the two distributions are, while the JS divergence fails. Therefore, our method will drive the agent to reach different areas as far as possible in the unknown space of valid states.
\begin{algorithm}
\caption{Naive WURL (test function)}
\label{alg:1}
\begin{algorithmic}
\State Initialize two policy $\pi_{\theta_{1}}$, $\pi_{\theta_{2}}$, and replay buffers for each policy $\mathcal{D}_{1}=\{\}$, $\mathcal{D}_{2}=\{\}$. Intialize test functions.
\While{Maximum number of episode is not reached}
    \State Select policy $l$ randomly or in turn. \texttt{done} = \texttt{False}.
    \While{Not \texttt{done}}
        \State Sample action from $\pi_{\theta_{l}}$, execute, and receive $s'$ and \texttt{done}.
        \If{$l=1$}
            \State Set reward $r=f(s)$ or $r=\mu(s)$.
        \Else 
            \State Set reward $r=-f(s)$ or $r=-\nu(s)$.
        \EndIf
        \State $\mathcal{D}_{l}$=$\mathcal{D}_{l}\cup\{(s,a,s',r)\}$.
        \State Train $\pi_{\theta_{l}}$ with SAC.
        \State Train test functions by sampling $\mathcal{D}_{1}$, $\mathcal{D}_{2}$.
    \EndWhile
\EndWhile
\end{algorithmic}
\end{algorithm}
\subsection{Wasserstein distance as intrinsic reward}
The Wasserstein distance can only measure discrepancy between two distributions. Therefore for the most naive approach to Wasserstein unsupervised reinforcement learning (WURL), we train a policy pair parameterized by $\pi_{\theta_{1}},\pi_{\theta_{2}}$, with Wasserstein distance between the state distributions $p_{\pi_{\theta_{1}}}(s),p_{\pi_{\theta_{2}}}(s)$ as their intrinsic reward.\par
As \citet{pacchiano2020learning} mentioned, dual form estimation allows us to assign reward at every step using test functions. We adopt two manners of dual formulation, TF1 (\cite{pmlr-v70-arjovsky17a}) and TF2 (\cite{abdullah2018reinforcement}). TF1 has one test function $f$ with Lipschitz constraint optimizing the objective in Eqn. \ref{eqn:2}.
Meanwhile TF2 has two test functions $\mu,\nu$ without any constraint. $\mu,\nu$ are trained according to Eqn. \ref{eqn:1}. As long as the test functions are optimal dual functions, the test functions give scores of each state. By splitting the maximization objective in Eqn. \ref{eqn:2}, we can assign $f(x)$ as reward for policy 1, and assign $-f(y)$ as reward for policy 2 to push $W_{c}(p,q)$ higher. Similar treatment can be applied to TF2. Combining RL training and test function training, we obtain Alg. \ref{alg:1}.\par
However, primal form estimation can only be executed after policy rollout by collecting states in one episode and compute the distance with states sampled from the replay buffer from another policy. We refer this pattern of reward granting to Alg. \ref{alg:2}. The challenge of sparse reward emerges in this training manner since the agent receives no reward until the episode ends. We will address this issue in Section 3.4.\par
Backend training algorithm of RL can vary. Off-policy algorithms like Soft-Actor-Critic (SAC) (\cite{haarnoja2018soft}) and on-policy algorithms like TRPO (\cite{schulman2015trust}), PPO (\cite{schulman2017proximal}) can all be deployed on WURL. Since SAC enjoys higher sample efficiency and suits environments with continuous action space, we choose SAC as our backend RL algorithm in our experiments.
\begin{algorithm}
\caption{Naive WURL (final reward)}
\label{alg:2}
\begin{algorithmic}
\State Initialize two policy $\pi_{\theta_{1}}$, $\pi_{\theta_{2}}$, and replay buffers for each policy $\mathcal{D}_{1}=\{\}$, $\mathcal{D}_{2}=\{\}$.
\While{Maximum number of episode is not reached}
    \State Select policy $l$ randomly or in turn. Set trajectory $\mathcal{S}=\{\}$. \texttt{done} = \texttt{False}.
    \While{Not \texttt{done}}
        \State Sample action from $\pi_{\theta_{l}}$, execute, and receive $s'$ and \texttt{done}.
        \State $\mathcal{S}=\mathcal{S}\cup\{s'\}$. Set reward $r=0$.
        \If{\texttt{done}}
            \State Sample target batch of states $\mathcal{T}$ from $\mathcal{D}_{3-l}$.
            \State Set reward $r=W(\mathcal{S},\mathcal{T})$ by any Wasserstein distance estimation method.
        \EndIf
        \State $\mathcal{D}_{l}$=$\mathcal{D}_{l}\cup\{(s,a,s',r)\}$.
        \State Train $\pi_{\theta_{l}}$ with SAC.
        \State Train test functions if WDE algorithm is one of the dual form methods.
    \EndWhile
\EndWhile
\end{algorithmic}
\end{algorithm}\par
\subsection{Learning with multiple policies}
Learning $N(N>2)$ policies at the same time requires the arbitrary policy $i$ to keep distance with all other policies. We expect the policy to maximize the average Wasserstein distance between state distribution $p_{i}$ induced by policy $i$, and other state distributions  $\frac{1}{N-1}\sum_{j=1,j\neq i}^{N}W(p_{i},p_{j})$. Also we can use the Wasserstein distance between $p_{i}$ and the average distribution of all others'. However this incurs underestimation due to the inequality
\begin{equation}
    W\left(p_{i},\frac{1}{N-1}\sum_{j=1,j\neq i}^{N}p_{j}\right)\leq\frac{1}{N-1}\sum_{j=1,j\neq i}^{N}W(p_{i},p_{j}).
\end{equation}
Practically we use $\min_{j=1,j\neq i}^{N}W(p_{i},p_{j})$ as reward to keep the current policy away from the nearest policy. As the number of policies growing, the number of times of distance computing grows as $\mathcal{O}(N^{2})$. The dual form requires $\mathcal{O}(N^{2})$ test functions which provokes memory and time consumption concern since every test function is a neural network and needs training. The primal form reward computation complexity also rises to $\mathcal{O}(N^{2})$. Fortunately, sliced Wasserstein distance or projected Wasserstein distance gives us a faster and more lightweight solution without training and inferring through neural networks.
\subsection{Amortized reward}
In contrast to test functions, the primal form estimation of Wasserstein distance produces one final reward at the end of an episode since we cannot estimate distribution distance from one sample. This nature of primal form estimation incurs sparse reward which imposes challenges on reinforcement learning and may impair the performance of value-based RL algorithms (\cite{NIPS2017_453fadbd}).\par
Noting that the primal form estimation automatically yields an optimal matching plan, we could decompose the overall Wasserstein distance into every sample. Formally speaking, suppose batch $\mathcal{S}=\{x_{n}\}^{N}_{n=1}$ is the set of states in one episode, and batch $\mathcal{T}=\{y_{m}\}^{M}_{m=1}$ is the state sample set of target distribution. We further suppose $P_{N\times M}$ is the optimal matching matrix given cost matrix $C_{N\times M}$. Denote $P_{i}$ as the $i$th row of matrix $P$, then the sample $x_{i}$ has its own credit by computing $P_{i}C^\top\mathbf{1}$ where $\mathbf{1}$ is $N\times 1$ vector filled with ones. Combining with projected Wasserstein distance, we have the following algorithm for crediting amortized reward. The matching matrix computation algorithm is stated in Appendix \ref{alg:a1}.
\begin{algorithm}
\caption{Amortized reward crediting}
\begin{algorithmic}
\State Given source batch $\mathcal{S}=\{x_{n}\}^{N}_{n=1}$ and target batch $\mathcal{T}=\{y_{m}\}^{M}_{m=1}$.
\State Compute cost matrix $C_{N\times M}$
\For{$k$ from 1 to $K$}
    \State Sample $v_{k}$ from $U(S^{d-1})$
    \State Compute projected samples $\hat{x}_{n}^{(k)}=\langle x_{n},v_{k}\rangle$, $\hat{y}_{m}^{(k)}=\langle y_{m},v_{k}\rangle$
    \State Compute matching matrix $P_{N\times M}^{(k)}$ from projected samples
    \State Compute reward vector $r^{(k)}=P^{(k)}C^\top\mathbf{1}$
\EndFor
\State Return mean reward vector $r=\frac{1}{K}\sum_{k=1}^{K}r^{(k)}$
\end{algorithmic}
\end{algorithm}

\begin{figure}[t]
    \centering
    \includegraphics[width=\linewidth]{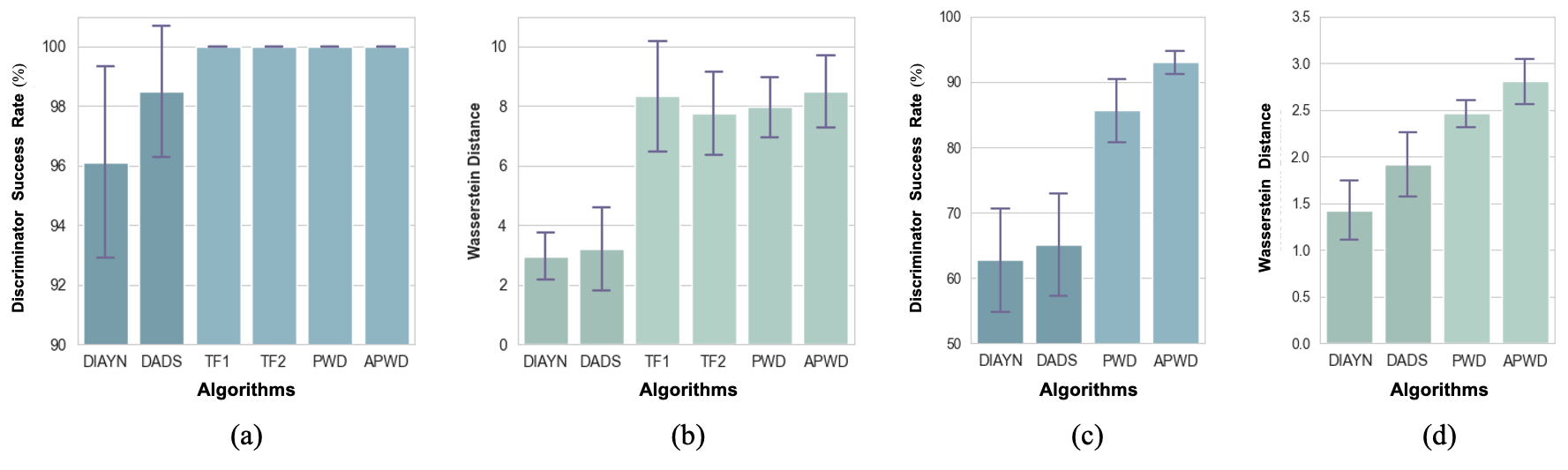}
    \caption{In the FreeRun environment, (a) and (b) show the results of learning 2 policies. (c) and (d) show the results of learning 10 policies at the same time. (a) and (c) show the success rate of the discriminator distinguishing policies. (b) and (d) show the mean Wasserstein distance between every two policies. Error bars represent standard deviations across 5 random runs.}
    \label{fig:exp_1}
\end{figure}

\subsection{Training schedule}
Our proposed method can either train $N$ policies at the same time or train incrementally by introducing new policies. Provided $N$ diverse policies, a new policy is trained to maximize the average Wasserstein distance from other policies by collecting state samples of policy $1,2,\dots,N$ at the beginning. The incremental training schedule provides flexibility of extending the number of diverse policies, especially when we are agnostic about how many policies are suitable for a certain environment beforehand. On the contrary, mutual information based unsupervised reinforcement learning is limited by fixed number of policies that cannot be easily extended due to the fixed neural network structure.\par
Similar idea appears in \citet{achiam2018variational}, in which skills are trained by a curriculum approach. The objective of curriculum training is to ease the difficulty of classifying large number of skills. However, the maximum number of skills is still fixed in advance therefore one cannot flexibly add new policies in.
\section{Experiments}
\subsection{Policy diversity}
We first examine our methods on a FreeRun environment where the particle agent spawns at the center. The agent can only control the acceleration on the X-Y plane and take the current position and velocity as observation. The particle runs freely on the plane with a velocity limit for a fixed number of steps. We compare the performance of the two groups of algorithms:

\begin{figure}[t]
    \centering
    \includegraphics[width=\linewidth]{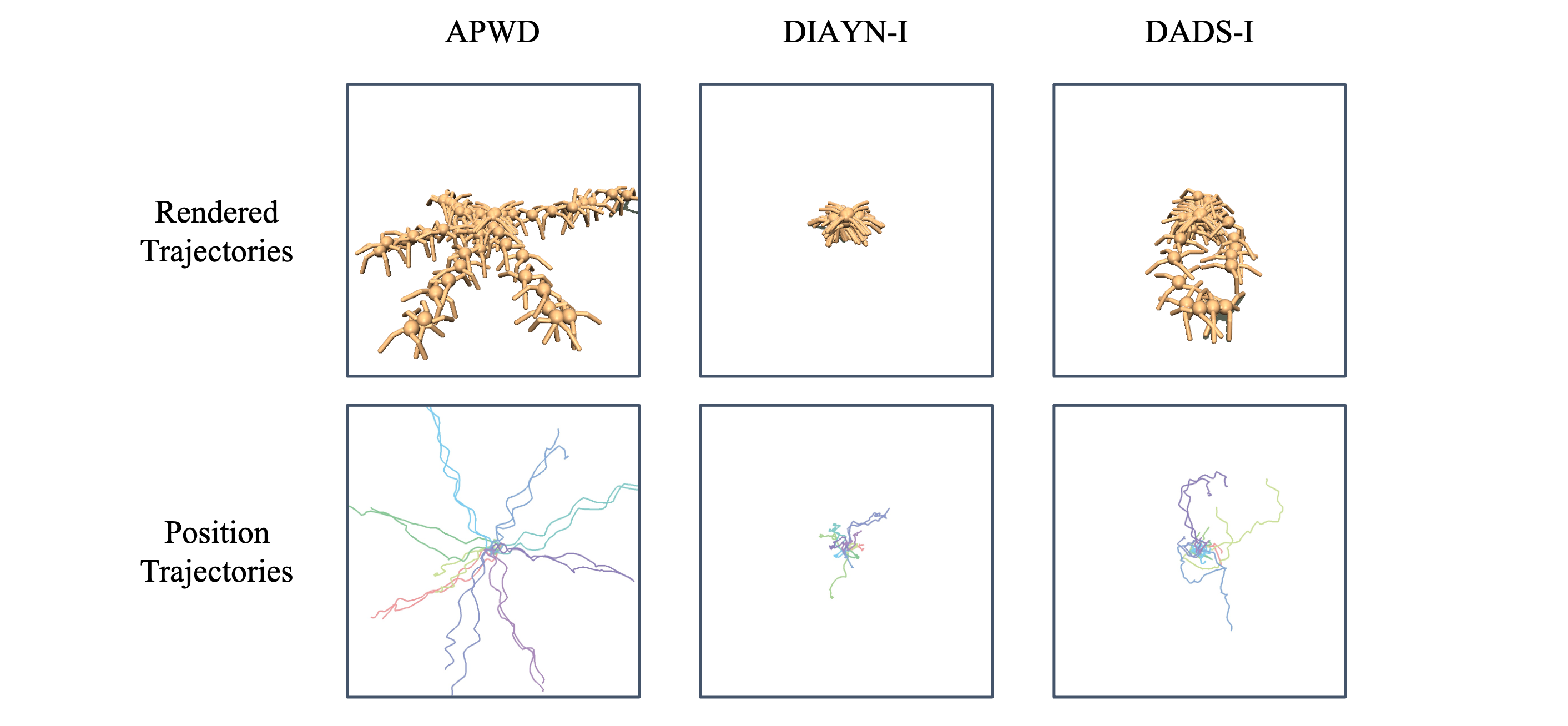}
    \caption{Visualized policies in MuJoCo Ant environment. The left column shows the rendered trajectories of 5 out of 10 total policies. The right column shows the X-Y position trajectories of 10 policies with different colors.}
    \label{fig:mujoco_vis}
\end{figure}

\begin{figure*}[t]
    \centering
    \includegraphics[width=0.9\textwidth]{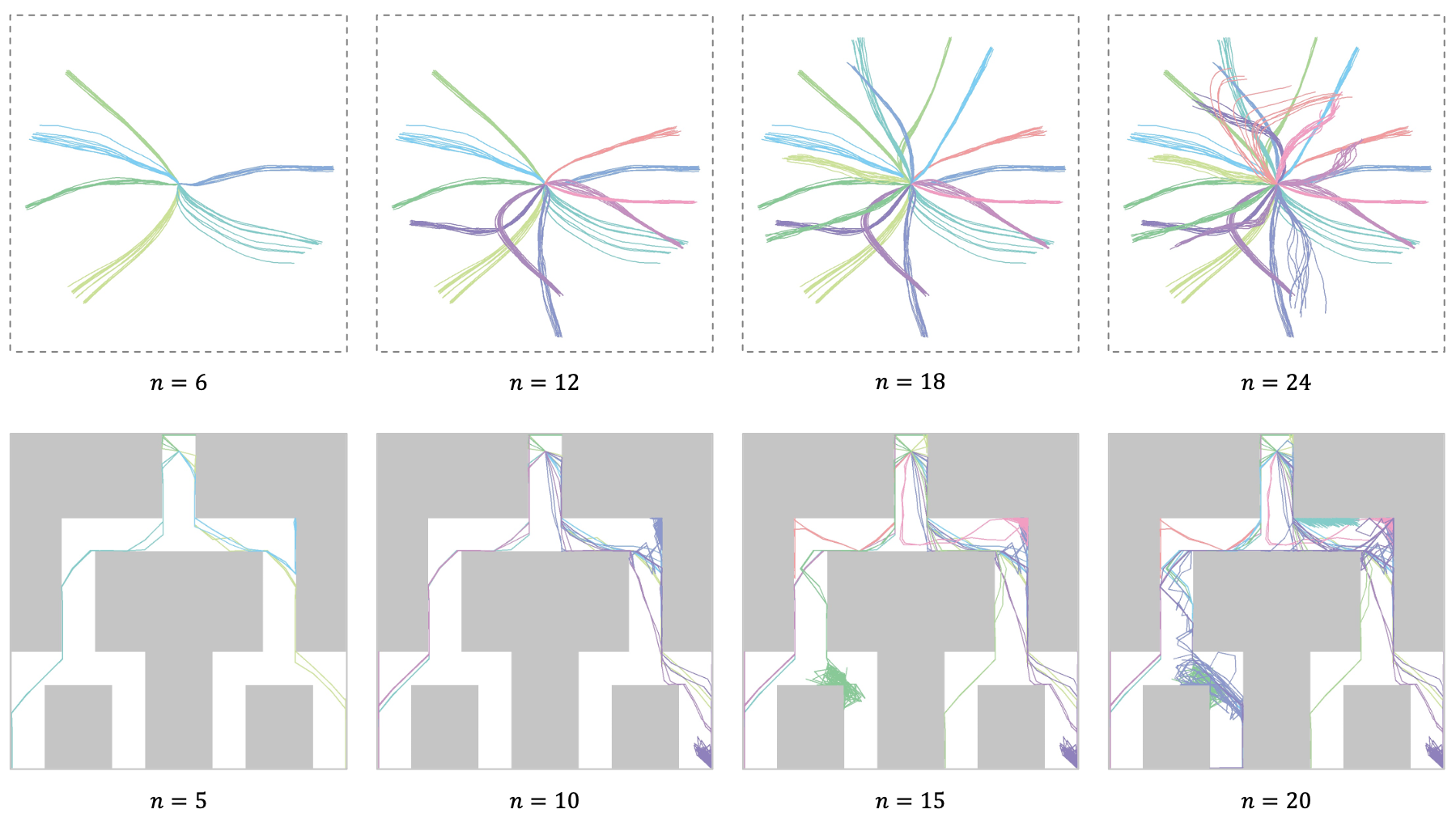}
    \caption{Results of training a bunch of policies incrementally. From left to right, new policies tend to reach new territory. As the number of policy $n$ grows, the policies gradually fill the state space.}
    \label{fig:exp_2}
\end{figure*}

\begin{table}[!t]
\centering
\begin{tabular}{ccccc}
\toprule   
Algorithm                   &   Metric  &   HalfCheetah                 &   Ant                         & Humanoid                      \\  
\midrule   
\multirow{2}*{APWD(Ours)}   &   DSR     &   0.981 $\pm$ 0.010           &   0.989 $\pm$ 0.002           &   0.998 $\pm$ 0.002           \\
~                           &   WD      &   \textbf{39.21 $\pm$ 0.44}   &   \textbf{8.57 $\pm$ 0.39}    &   \textbf{556.96 $\pm$ 27.78} \\
\specialrule{0em}{2pt}{2pt}
\multirow{2}*{DIAYN-I}      &   DSR     &   0.978 $\pm$ 0.006           &   0.985 $\pm$ 0.001           &   0.999 $\pm$ 0.001           \\
~                           &   WD      &   13.01 $\pm$ 2.89            &   4.50 $\pm$ 0.27             &   279.94 $\pm$ 27.98          \\
\specialrule{0em}{2pt}{2pt}
\multirow{2}*{DADS-I}       &   DSR     &   0.994 $\pm$ 0.003           &   0.793 $\pm$ 0.035           &   0.999 $\pm$ 0.001           \\
~                           &   WD      &   11.75 $\pm$ 1.90            &   7.53 $\pm$ 0.27             &   420.67 $\pm$ 56.83          \\
\bottomrule  
\end{tabular}
\caption{Policy diversity on three MuJoCo locomotion environments. Metric DSR represents the discriminator success rate and WD denotes the mean Wasserstein distance between two policies.}
\label{tab1}
\end{table}

\begin{itemize}
    \item Mutual information as intrinsic reward: DIAYN, DADS;
    \item Wasserstein distance as intrinsic reward: TF1, TF2, PWD, APWD.
\end{itemize}
DADS (\cite{sharma2019dynamics}) is another mutual information based algorithm that maximizes $I(S_{t+1};Z|S_{t})$ utilizing model-based framework. It claims lower entropy of learned skills than DIAYN. TF1 and TF2 are two methods which adopt test function optimizing Eqn. \ref{eqn:2} and Eqn. \ref{eqn:1} respectively. PWD (Projected Wasserstein Distance) uses a primal form Wasserstein distance estimation method described in Section 2.2, and APWD is the amortized version of PWD to avoid sparse reward stated in Section 3.4. In the setting of learning 10 policies, TF1 and TF2 are unavailable since they are not compatible with multiple policies as mentioned in Section 3.2. We examine these methods from two aspects. First, we train a neural network (discriminator) to distinguish each policy from rollout state samples, and compare the success rate of the discriminator. Second, we estimate the mean Wasserstein distance between every two state distributions of two different policies. Fig. \ref{fig:exp_1} shows policies learned by Wasserstein distance based algorithms generally has greater discriminability, and larger distance between the state distributions of every two policies. Fig. \ref{fig:exp_1} also demonstrates amortized reward improves performance.\par

We also examine our algorithms on several MuJoCo tasks, including three classical locomotion environments, and two customized point mass environments where a simplified ball agent wanders in different maps with various landscapes and movable objects (see Appendix for demonstrations). Considering the significantly larger action space and state space in MuJoCo environments, we replace the shared actor network $\pi(a|s,z),z=\texttt{OneHot}(i)$ in DIAYN and DADS with $N$ individual networks $\pi_{i}(a|s),i=1,\dots,N$, while keep the discriminator network unchanged. DIAYN and DADS with individual actors (DIAYN-I and DADS-I for short) enjoy greater parameter space and they are capable to learn a more diverse set of policies in MuJoCo tasks. Table \ref{tab1} shows in the most cases, all three unsupervised RL approaches yield highly distinguishable policies. However, APWD achieves better performance on Wasserstein distance, which means the policies learned by APWD are more distantly distributed in the state space. Fig. \ref{fig:mujoco_vis} visualizes the differences of three policy sets in MuJoCo Ant environment and clearly shows that APWD encourages the policies to keep far from each other. The results verify our hypothesis in Section 3.1, that mutual information based intrinsic reward is unable to drive the policies far from each other when JS divergence saturates.
\subsection{Incremental learning}
Our method provides flexibility to enlarge the policy set. New policies can be trained one by one, or trained based on policies in hand. We show the process of incremental learning to illustrate how the newly learned policies gradually fill the state space in Fig. \ref{fig:exp_2}. As new policies added in, the particle agent in FreeRun and TreeMaze runs to new directions and new areas, and behaves differently (dithering around, turning, etc.).

\begin{table}[!t]
\centering
\begin{tabular}{ccccc}
\toprule   
Algorithm   &   FreeRun                         &   Ant                             \\  
\midrule   
APWD(Ours)  &   \textbf{125.56 $\pm$ 12.63}     &   \textbf{100.00 $\pm$ 18.03}     \\
\specialrule{0em}{2pt}{2pt}
DIAYN-I     &   15.06 $\pm$ 26.97               &   35.00 $\pm$ 7.07                \\
\specialrule{0em}{2pt}{2pt}
DADS-I      &   109.78 $\pm$ 27.08              &   46.00 $\pm$ 10.84                \\
\bottomrule  
\end{tabular}
\caption{Rewards of meta-policies on two hierarchical RL scenarios. Each meta-policy is trained with a sub-policy set which contains 10 policies pre-trained with a specific unsupervised RL algorithm listed above.}
\label{tab2}
\end{table}

\subsection{Downstream tasks}
In previous unsupervised RL literature, the policies (or skills) learned without any task reward can be utilized either in hierarchical reinforcement learning or in planning (\cite{eysenbach2018diversity}; \cite{sharma2019dynamics}). Likewise, we examined our methods on downstream tasks including two navigation tasks based on the particle environment FreeRun and MuJoCo Ant. Both tasks require the agent to reach specific goals in a given order. The agent receives +50 reward for each goal reached. In FreeRun navigation task, we penalize each step with small negative reward to encourage the agent to finish the task as quickly as possible.\par
To tackle these navigation tasks with pre-trained policies, we employ a meta-policy to choose one sub-policy to execute during $H$ steps ($H$ is fixed in advance in our tasks). The meta-policy observes agent state every $H$ steps , chooses an action corresponded to a sub-policy, and then receives the reward that the sub-policy collected during the successive $H$ steps. Therefore, we can train the meta-policy with any compatible reinforcement learning algorithms. In our experiments, we adopt PPO as meta-policy trainer (\cite{schulman2017proximal}).\par
Table \ref{tab2} presents the results on FreeRun and MuJoCo Ant navigation tasks. The pre-trained 10 policies with DIAYN-I, DADS-I and our proposed method APWD serve as the sub-policies in the hierarchical framework. Since our method yields more diverse and distant sub-policies, APWD based hierarchical policy achieves higher reward in both navigation tasks without doubt.
\section{Related work}
Learning in a reward-free environment has been attracting reinforcement learning researchers for long. Early research takes mutual information as maximization objective. They explored the topics on which variable of the policy should be controlled by the latent variable, and how to generate the distribution of the latent variable. VIC (\cite{gregor2016variational}) maximizes $I(Z;S_{f})$ to let the final state of a trajectory be controlled by latent code $Z$, while allowing to learn the prior distribution $p(z)$. VALOR (\cite{achiam2018variational}) takes similar approach maximizing $I(Z;\tau)$ where $\tau$ denotes the whole trajectory, but keeps $p(z)$ fixed by Gaussian distribution. \citet{hausman2018learning} uses a network to embed various tasks to the latent space. DIAYN (\cite{eysenbach2018diversity}) improves the performance by maximizing $I(Z;S)$, fixing prior distribution, while minimizes $I(Z;A|S)$. Recent papers show their interests on better adapting unsupervised skill discovery algorithm with MDP, on the aspect of transition model and initial state. DADS (\cite{sharma2019dynamics}) gives a model based approach by maximizing $I(S';Z|S)$ so that the learned skills can be employed in planning. \citet{baumli2020relative} alternates the objective to $I(S_{f};Z|S_{0})$ in order to avoid state partitioning skills in case of various start states.\par
However, the nature of these methods restrict the diversity of learned skills since the mutual information is upper bounded by $H(Z)$. Recent research on mutual information based methods tries to enlarge the entropy of the prior distribution through fitting the uniform distribution on valid state space $U(S)$. EDL (\cite{campos2020explore}) takes three seperate steps by exploring the state space, encoding skills and learning skills. EDL first uses state marginal matching algorithm to yield a suffiently diverse distribution of states $p(s)$, then a VQ-VAE is deployed to encode the state space to the latent space, which creates $p(z|s)$ as the discriminator to train the agent. Skew-fit (\cite{pong2019skew}) adopts goal conditioned policies where the goal is sampled through importance sampling in the skewed distribution of $p(s)$ acquired by current policy. The agent can gradually extend their knowledge of state space by fitting the skewed distribution. Both methods claim they have state-covering skills.\par
Wasserstein distance as an alternative distribution discrepancy measure is attracting machine learning researchers recently (\cite{ozair2019wasserstein}). Especially in the literature of generative models (\cite{pmlr-v70-arjovsky17a}; \cite{NEURIPS2018_2c89109d}; \cite{patrini2020sinkhorn}; \cite{tolstikhin2018wasserstein}), Wasserstein distance behaves well in the situations where distributions are degenerate on a sub-manifold in pixel space. In reinforcement learning, Wasserstein distance is used to characterize the differences between policies instead of commonly used $f$-divergences, e.g., KL divergence (\cite{pmlr-v80-zhang18a}). \citet{pacchiano2020learning} reports improvements in trust region policy optimization and evolution strategies, and \citet{dadashi2021primal} shows its efficacy in imitation learning by minimizing Wasserstein distance between behavioral policy and expert policy. Our proposed method inherits the motivations of using Wasserstein distance as a new distribution discrepancy measure in generative models and policy optimization. To increase the diversity of policies in unsupervised reinforcement learning, Wasserstein distance appears to be a more appropriate measure than $f$-divergences derived from mutual information based methods.
\section{Discussion}
\subsection{Limitations}
Our work uses Wasserstein distance as a new metric measuring discrepancy between probability distributions. Although this new metric provides better performance on unsupervised reinforcement learning in certain environments stated in Section 4, there are limitations or difficulties for further application. First, Wasserstein distance depends on cost function $c(x,y)$. In our experiments, we choose $l_{2}$-norm since the agent is conducting navigation tasks. However, choosing a proper cost function may be difficult for other reinforcement learning environments. Second, different dimensions of state space may have different semantic intention such as position, force, angle. Therefore, implementing WURL in the full dimension of state space may not properly balance between different dimensions. Third, image-based observations could not be used for calculating Wasserstein distance directly. Nevertheless, these limitations or difficulties in deploying WURL in a larger range of applications may imply future research directions.\par
\subsection{Conclusion}
We build a framework of Wasserstein unsupervised reinforcement learning (WURL) of training a set of diverse policies. In contrast to conventional methods of unsupervised skill discovery, WURL employs Wasserstein distance based intrinsic reward to enhance the distance between different policies, which has theoretical advantages to mutual information based methods (\cite{pmlr-v70-arjovsky17a}). We overcome the difficulties of extending the WURL framework for multiple policy learning. In addition, we devise a novel algorithm combining Wasserstein distance estimation and reinforcement learning, addressing reward crediting issue. Our experiments demonstrate WURL generates more diverse policies than mutual information based methods such as DIAYN and DADS, on the metric of discriminability (MI-based metric) and Wasserstein distance. Furthermore, WURL excites autonomous agents to form a set of policies to cover the state space spontaneously and provides a good sub-policy set for sequential navigation tasks.
{
\bibliographystyle{plainnat}
\bibliography{main}
}
\appendix

\section{Appendix}
\subsection{Evaluation of different estimation methods of Wasserstein distance}
\begin{figure}[h]
    \centering
    \includegraphics[width=\textwidth]{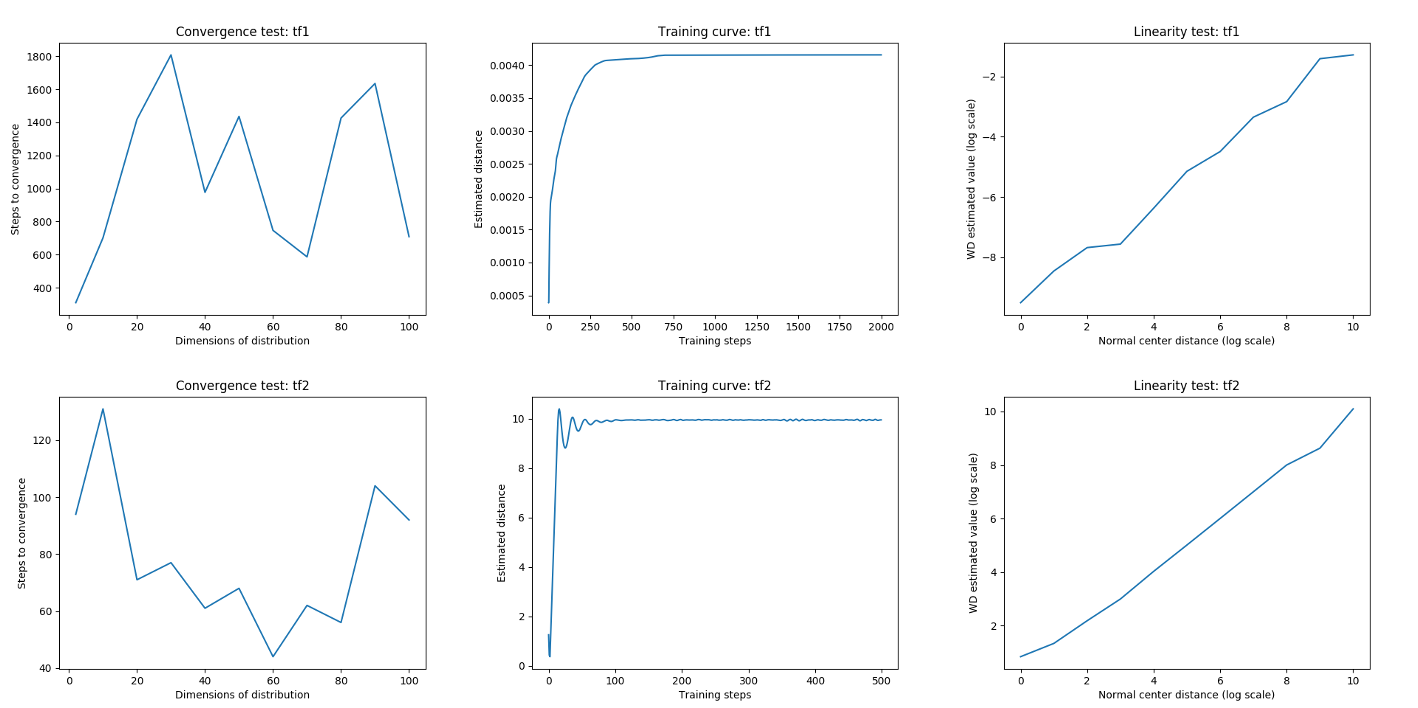}
    \caption{Comparisons of TF1 and TF2, on the aspects of iterations to convergence, typical training curve and linearity.}
    \label{fig:tf}
\end{figure}

In this section we will compare different estimation methods of Wasserstein distance, including two dual form methods TF1, TF2 and two primal form methods SWD, PWD. As we mentioned in main text, TF1 maximizes dual objective with one test function $f$ subject to Lipschitz constraint. In practice, the Lipschitz constraint is implemented by clamping weights in $f$ parameterized by a neural network. We set this constant to 0.01. TF1 cannot give the exact number Wasserstein distance but the distance multiplied by some underlying constant. Therefore, TF1 is still capable of measuring how ``far'' two distributions are. TF2 maximizes dual objective plus a regularizer term with two test functions $\mu,\nu$ subject to no constraint. TF2 can provide the exact number of Wasserstein distance. Fig. \ref{fig:tf} shows the typical training curves, convergence speed versus data dimension, and the linearity as the distribution distance grows. All experiments are evaluated on two set of samples sampled from two different Gaussian distributions.\par

Since SWD projects samples on a real line and the distances are also projected, SWD tends to underestimate the true Wasserstein distance. Table \ref{tab:1} characterizes the estimated numbers of each method. As we can see, TF2 and PWD give exact number of Wasserstein distance, and PWD consume less time of execution than TF2. Considering time consumption and accuracy, PWD is preferred in our algorithm.
\begin{table}[t]
\centering
    \begin{tabular}{c|c|c}
        \hline
         & Distance estimated & Execution time (s) \\
         \hline
        Ground truth & 2.0 & N/A\\
        TF1 & 0.00341$\pm$0.00045 & 4.92$\pm$0.04\\
        TF2 & 2.92$\pm$0.05 & 1.43$\pm$0.05\\
        SWD & 0.791$\pm$0.026 & 0.00409$\pm$0.00003\\
        PWD & 3.21$\pm$0.002 & 0.00643$\pm$0.00003\\
        \hline
        Ground truth & 16.0 & N/A\\
        TF1 & 0.0135$\pm$0.0022 & 4.98$\pm$0.05\\
        TF2 & 16.1$\pm$0.0 & 1.39$\pm$0.02\\
        SWD & 6.31$\pm$0.22 & 0.00409$\pm$0.00002\\
        PWD & 16.3$\pm$0.0 & 0.00647$\pm$0.00004\\
        \hline
        Ground truth & 64.0 & N/A\\
        TF1 & 0.0349$\pm$0.0053 & 4.90$\pm$0.04\\
        TF2 & 64.1$\pm$0.0 & 1.40$\pm$0.03\\
        SWD & 25.9$\pm$0.75 & 0.00410$\pm$0.00007\\
        PWD & 64.2$\pm$0.0 & 0.00656$\pm$0.00015\\
        \hline
    \end{tabular}
    \vspace{0.25cm}
    \caption{Comparisons of distance estimation and execution time}
    \label{tab:1}
\end{table}
\subsection{Matching matrix computation}
Assume two 1-D (one dimensional) sample sets $\mathcal{X}=\{x_{i}\}_{i=1}^{N}$ and $\mathcal{Y}=\{y_{j}\}_{j=1}^{N}$. Further we assume the cost function is $C(x,y)=|x-y|$. First we sort the two sets to ordered sets $\{x_{\sigma_{x}(i)}\}_{i=1}^{N}$ and $\{y_{\sigma_{y}(j)}\}_{j=1}^{N}$. The analytical solution of optimal matching in Wasserstein distance estimation is given by $(x_{\sigma_{x}(i)},y_{\sigma_{y}(i)})$. Exact Wasserstein distance between the two empirical distributions induced by sample sets is
\begin{equation}
    W(P_{x},P_{y})=\frac{1}{N}\sum_{i=1}^{N}|x_{\sigma_{x}(i)}-y_{\sigma_{y}(i)}|=\frac{1}{N}\sum_{i=1}^{N}|x_{i}-y_{\sigma_{y}^{-1}\sigma_{x}(i)}|.
\end{equation}
The matching matrix is instantly acquired from permutations calculated above.\par
When two samples sets have different cardinalities, e.g., $\mathcal{X}=\{x_{i}\}_{i=1}^{N}$ and $\mathcal{Y}=\{y_{j}\}_{j=1}^{M}$. We duplicate the elements in $\mathcal{X}$ by $M$ times and duplicate the elements in $\mathcal{Y}$ by $N$ times. Then the two sets have the same cardinality $N\times M$, and the aforementioned algorithm can be used. Nevertheless, for computational convenience, we compute matching matrix in the following algorithm
\begin{algorithm}
\label{alg:a1}
\caption{Matching matrix computation}
\begin{algorithmic}
\State Sort two sets to $\{x_{\sigma_{x}(i)}\}_{i=1}^{N}$ and $\{y_{\sigma_{y}(j)}\}_{j=1}^{M}$
\State Initialize matching matrix $P_{N\times M}$ with zeros
\State Initialize list $\texttt{A}=[(1/N,\sigma_{x}(i))]_{i=1}^{N}$ and list $\texttt{B}=[(1/M,\sigma_{y}(j))]_{j=1}^{M}$
\State Set $u=0,v=0$
\While{$\texttt{A}\neq\emptyset$ and $\texttt{B}\neq\emptyset$}
    \State $u,k=\texttt{A.pop()}$ if $u=0$, $v,l=\texttt{B.pop()}$ if $v=0$
    \State $w=min(u,v)$
    \State $P_{kl}+=w$
    \State $u-=w$, $v-=w$
\EndWhile
\end{algorithmic}
\end{algorithm}
\subsection{Experiment settings}
Hardware:
\begin{itemize}
    \item 1x GeForce RTX 2080,
    \item 1x Intel Core i7-7700 CPU @ 3.60GHz.
\end{itemize}
Software:
\begin{itemize}
    \item Ubuntu 18.04.3 LTS,
    \item torch 1.7.1+cu102,
    \item python 3.7.5,
    \item MuJoCo 150,
    \item Gym 0.18.0.
\end{itemize}
Network and optimizer:
\begin{itemize}
    \item Network type: MLP,
    \item hidden layer size: 64 for environments in mujoco-maze, 256 for HalfCheetah, Ant and 1024 for Humanoid,
    \item activation layer: ReLU,
    \item optimizer: Adam.
\end{itemize}
Please refer to source code for other details. Part of environments are customized, i.e., FreeRun, TreeMaze. Part of environments are modified from mujoco-maze 0.1.1, i.e., PointPush, PointBilliard. The modified environments are presented in source code.
\subsection{Additional experiments}
\subsubsection{Policies trained in MuJoCo locomotion environments}
\begin{figure}[ht]
    \centering
    \includegraphics[width=\textwidth]{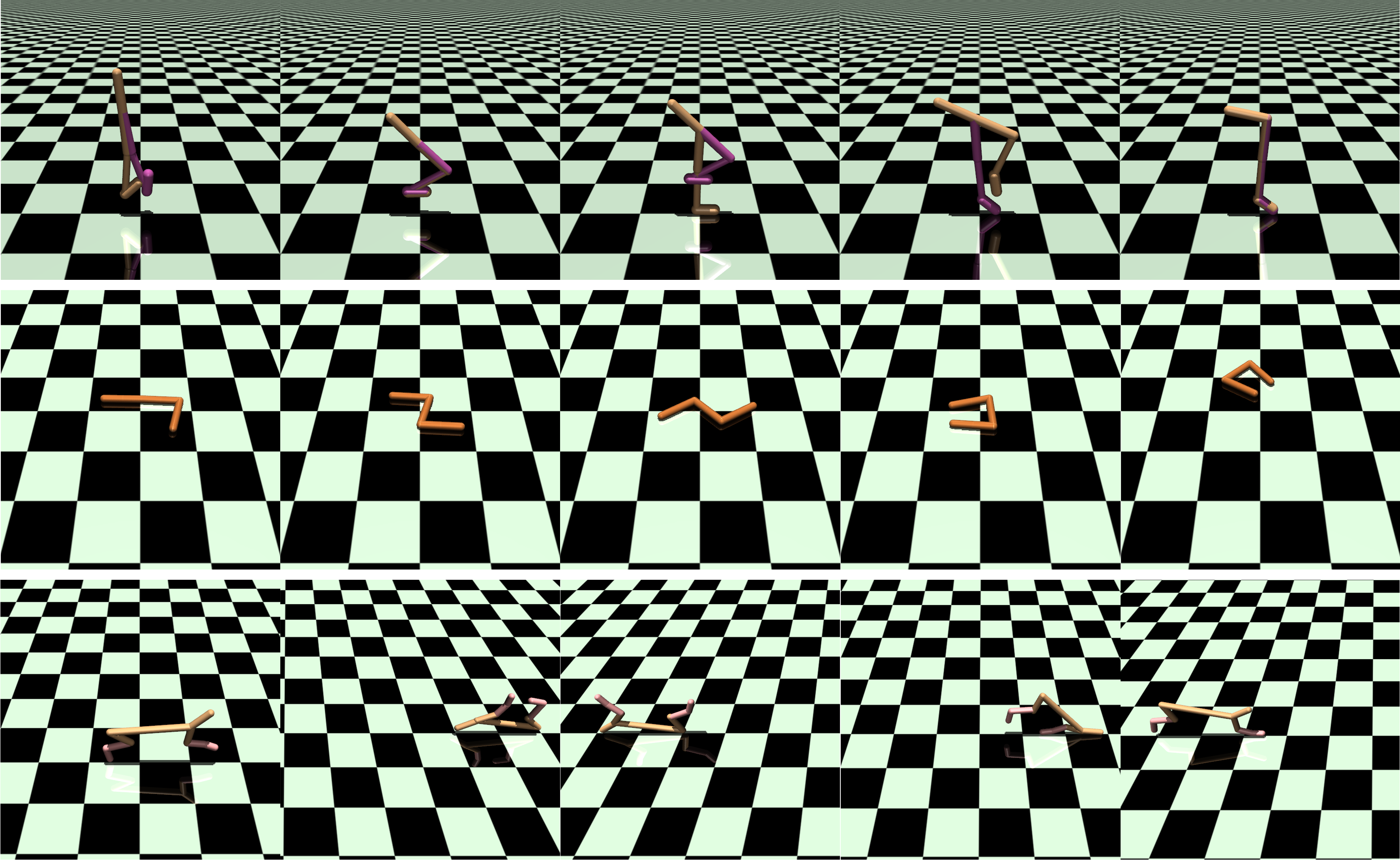}
    \caption{Policies learned in Walker2d, Swimmer, HalfCheetah.}
    \label{fig:a3}
\end{figure}

We also investigate our proposed method in MuJoCo locomotion tasks, e.g., Walker2d, Swimmer, HalfCheetah. As Fig. \ref{fig:a3} shows, under different policies, the agent acts with different gestures, or moves to different directions. These trained models are provided in codes.
\subsubsection{Minimum distance as reward vs. mean distance as reward}
\begin{figure}[ht]
    \centering
    \includegraphics[width=0.6\textwidth]{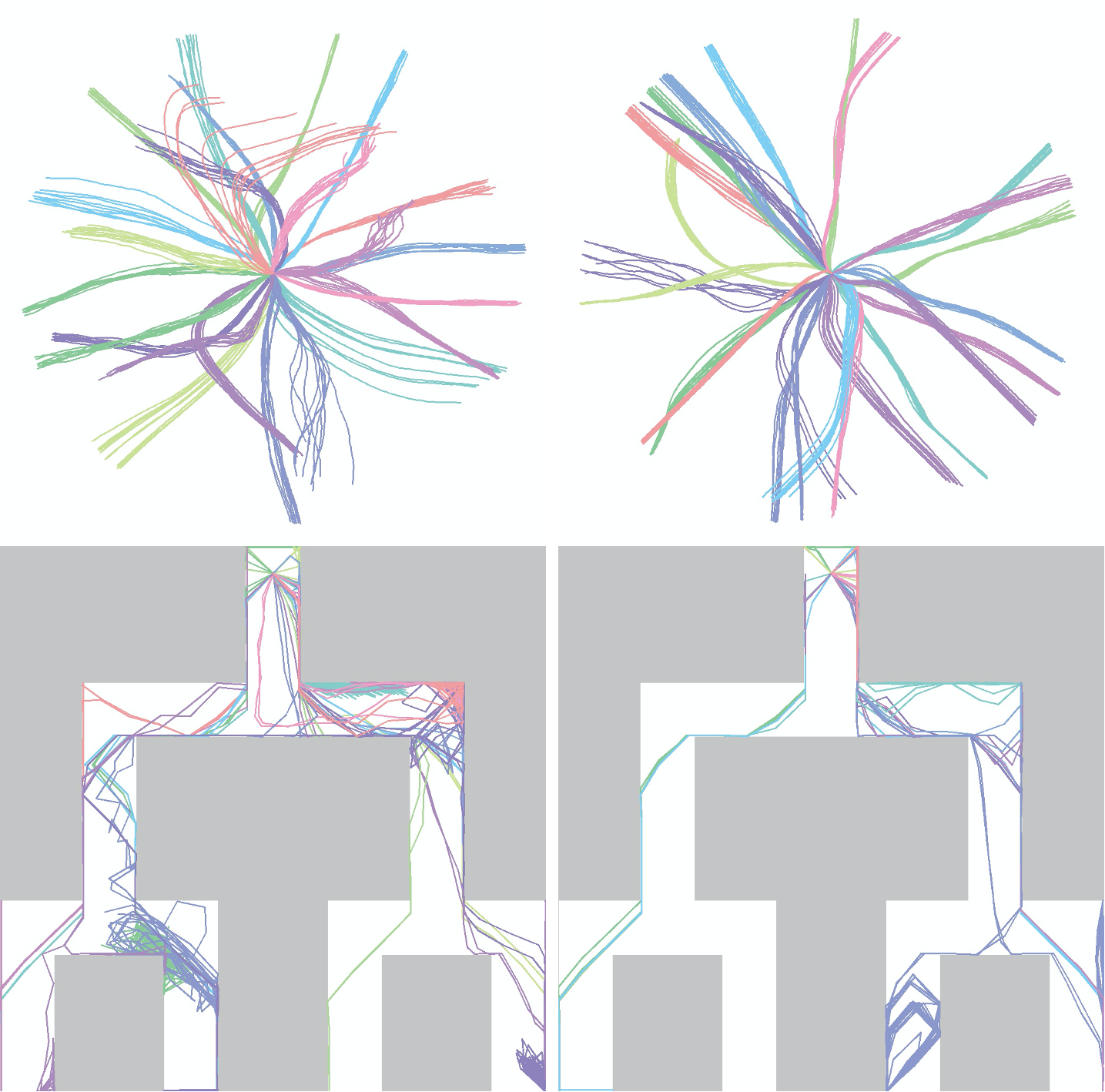}
    \caption{Left column: minimum distance as reward; right column: mean distance as reward. 24 policies learned in FreeRun and TreeMaze environments.}
    \label{fig:a1}
\end{figure}

In Section 3.3, we mentioned we use the minimum distance as reward $\min_{j=1,j\neq i}^{N}W(p_{i},p_{j})$ in practical use, not the mean distance $\frac{1}{N-1}\sum_{j=1,j\neq i}^{N}W(p_{i},p_{j})$. We compare this two methods and visualize the learned trajectories in Fig. \ref{fig:a1}. Maximizing the minimum distance can lead the current policy away from any other policies however maximizing the mean distance results in similar policies as demonstrated in the right column of Fig. \ref{fig:a1}.
\subsubsection{Trajectories of models from DIAYN and APWD}
\begin{figure}[ht]
    \centering
    \includegraphics[width=0.6\textwidth]{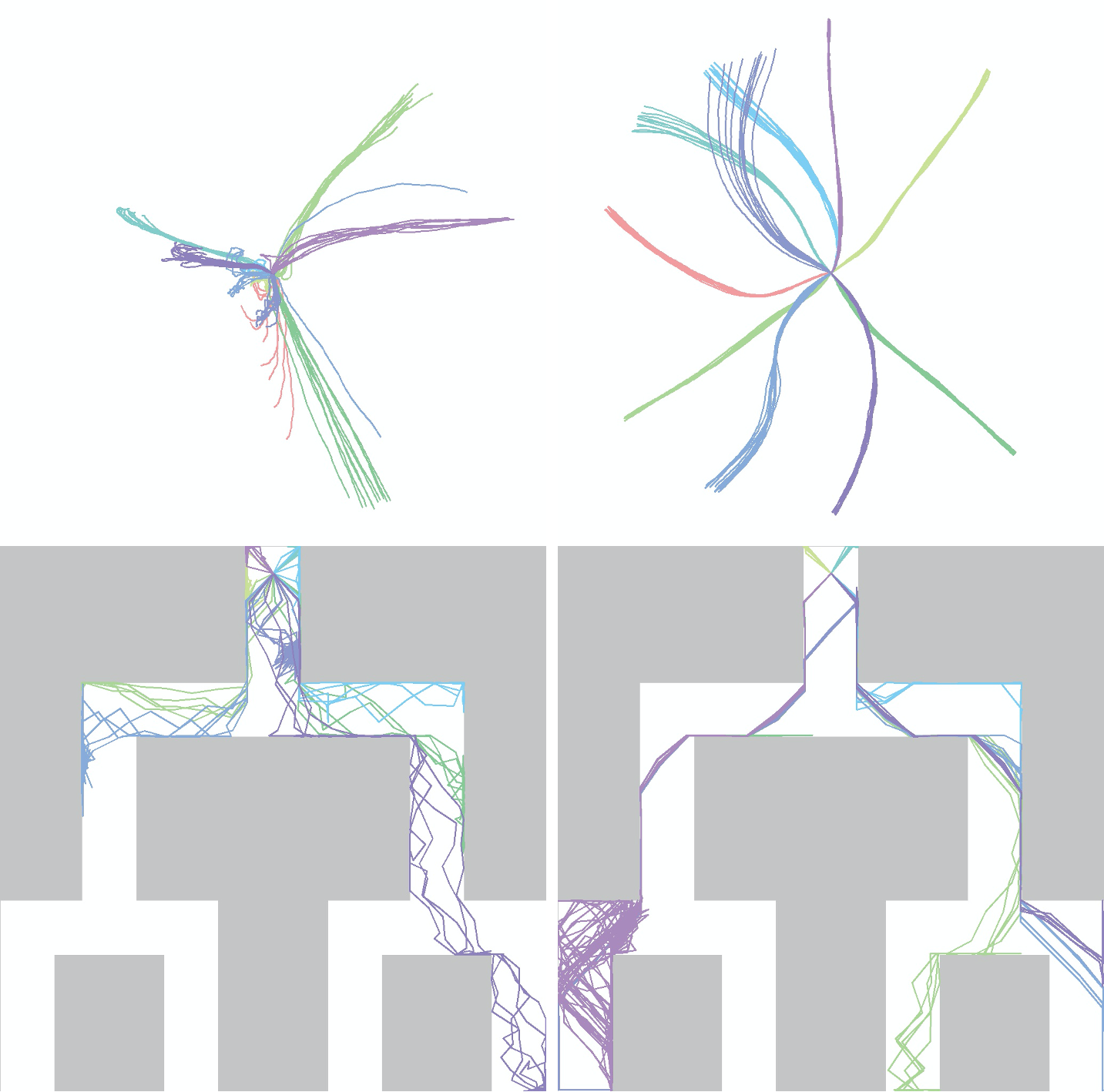}
    \caption{Left column: DIYAN; right column: WURL with amortized PWD. 10 policies learned in FreeRun and TreeMaze environments.}
    \label{fig:a2}
\end{figure}

We visualize the policies learned by DIAYN and APWD. As we can see in Fig. \ref{fig:a2}, maximizing Wasserstein distance keeps the policies far from each other. Although policies learned by DIAYN can be distinguished by a discriminator, they would not keep far and explore the outer state space spontaneously.

\subsubsection{Results on customized MuJoCo environments}
\begin{figure}[ht]
    \centering
    \includegraphics[width=0.95\textwidth]{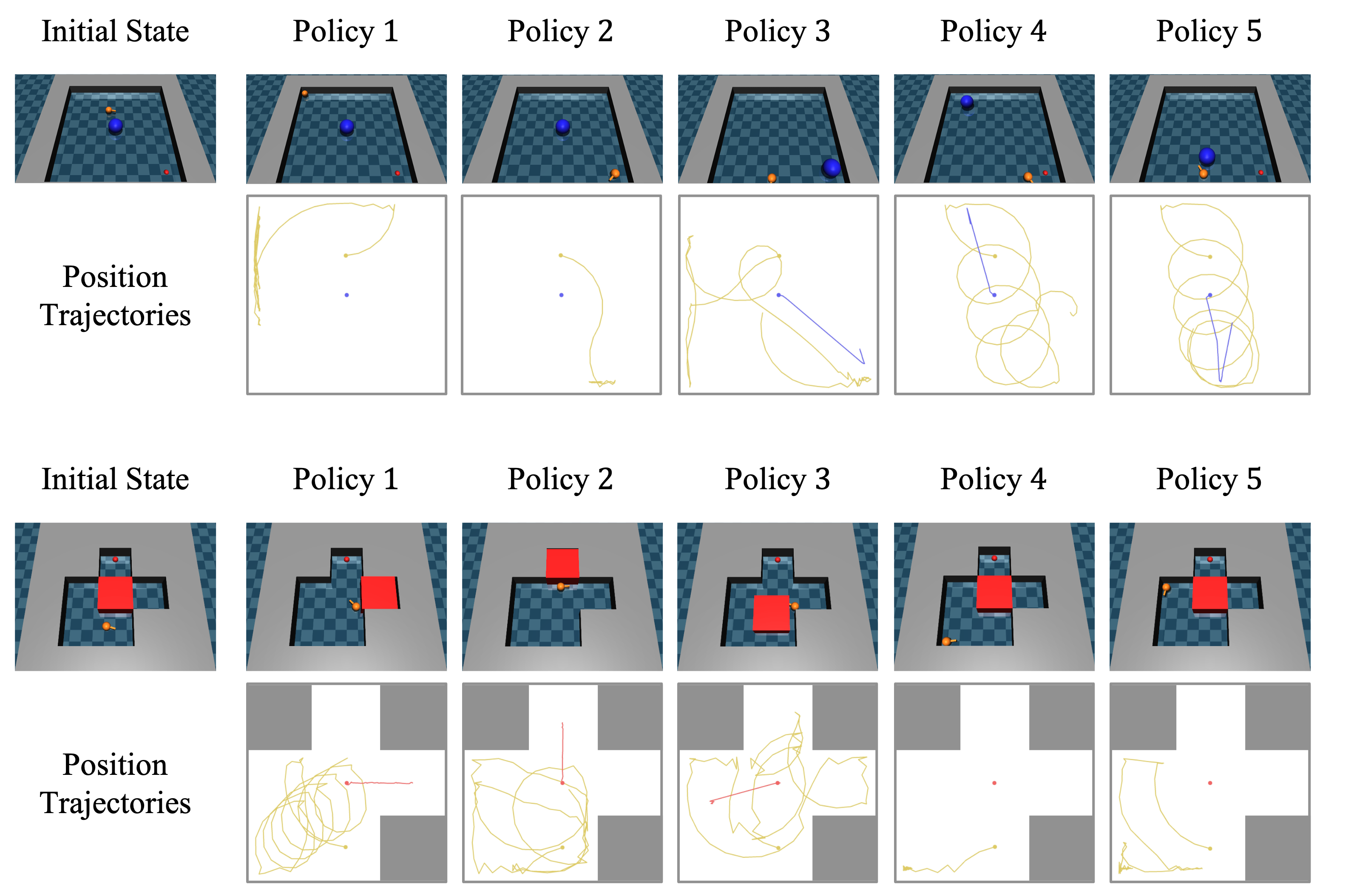}
    \caption{5 Policies learned in customized MuJoCo environments, PointBilliard (upper) and PointPush (bottom). We visualize the position trajectories of the centroid of movable objects as well which demonstrate the differences of the learned policies.}
    \label{fig:a_point_maze}
\end{figure}

As mentioned in main text Section 4.1, we applied WURL on two customized MuJoCo environments, PointPush and PointBilliard. Fig. \ref{fig:a_point_maze} shows selected policies learned by APWD algorithm. Not only the agent moves to different corner on the map, but the agent learns to interact with the movable object as well, since this behaviour will enhance the diversity.

\subsubsection{Training details of hierarchical reinforcement learning experiments}
\begin{figure}[ht]
    \centering
    \includegraphics[width=0.8\textwidth]{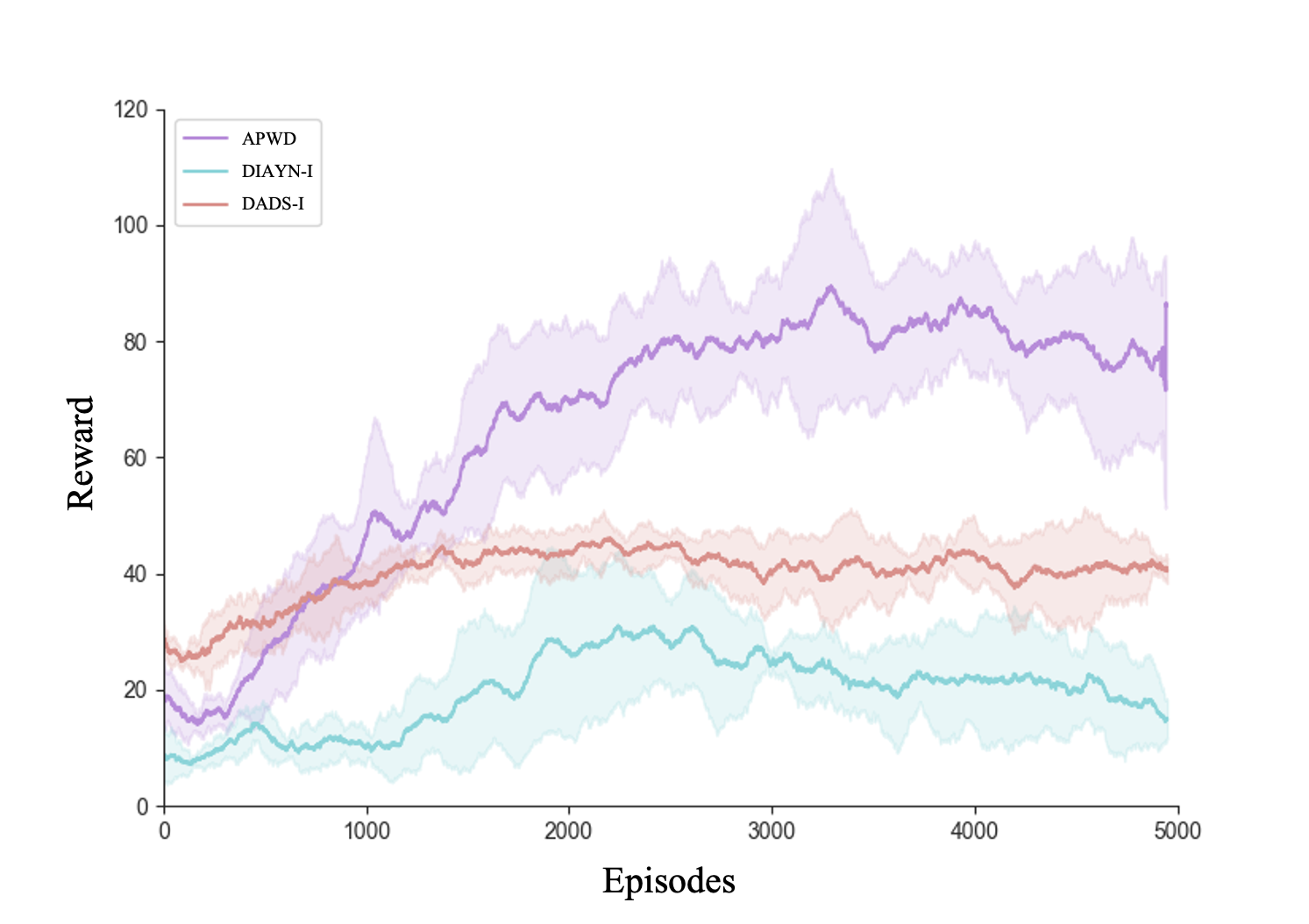}
    \caption{The training curve of the navigation task based on MuJoCo Ant environment. The solid line is the average return across 5 random runs and the shadowed area denotes the standard deviation.}
    \label{fig:a_hrl_ant_curve}
\end{figure}

We compare the performance of our method APWD with DIAYN-I and DADS-I in downstream tasks as mentioned in main text Section 4.3. The 10 pre-trained policies from each method serve as the sub-policies in the hierarchical framework and we adopt PPO as the meta-controller trainer. Fig. \ref{fig:a_hrl_ant_curve} is the training curve of the navigation task based on MuJoCo Ant environment and shows that the pre-trained policies from APWD outperform policies from the other two methods in this hierarchical framework.

\end{document}